%% file: aaai24.tex
\documentclass[letterpaper]{article} 
\usepackage{aaai24}  
\usepackage{times}  
\usepackage{helvet}  
\usepackage{courier}  
\usepackage[hyphens]{url}  
\usepackage{graphicx} 
\usepackage{natbib}  
\usepackage{caption} 
\usepackage{algorithm}
\usepackage{algorithmic}

\usepackage{amssymb}
\usepackage{subcaption}
\usepackage{xcolor}
\usepackage{amsmath}
\usepackage{hyperref}
\usepackage{newfloat}
\usepackage{listings}
\DeclareCaptionStyle{ruled}{labelfont=normalfont,labelsep=colon,strut=off} 
\floatstyle{ruled}
\newfloat{listing}{tb}{lst}{}
\floatname{listing}{Listing}
\title{Critic-Guided Decision Transformer for Offline Reinforcement Learning}
\author {
    Yuanfu Wang\thanks{Equal contribution.}\textsuperscript{\rm 1, \rm 2},
    Chao Yang\footnotemark[1]\textsuperscript{\rm 2},
    Ying Wen\thanks{Correspondence to Ying Wen, Yu Qiao.}\textsuperscript{\rm 1},
    Yu Liu\textsuperscript{\rm 2, \rm 3},
    Yu Qiao\footnotemark[2]\textsuperscript{\rm 2}
}
\affiliations {
    \textsuperscript{\rm 1}Shanghai Jiao Tong University\\
    \textsuperscript{\rm 2}Shanghai Artificial Intelligence Laboratory\\
    \textsuperscript{\rm 3}SenseTime Research\\
    \{wangyuanfu,yangchao,qiaoyu\}@pjlab.org.cn, ying.wen@sjtu.edu.cn, liuyuisanai@gmail.com
}

\begin{document}

\maketitle

\input{sections/0_abs}

\input{sections/1_intro}

\input{sections/2_related}

\input{sections/3_problem}

\input{sections/4_method}
\input{sections/5_exp}

\input{sections/6_conc}

\section*{Acknowledgments}
This research is funded by the Shanghai Artificial Intelligence Laboratory. It also received partial support from the Shanghai Post-doctoral Excellent Program (Grant No. 2022234). Additionally, the team from Shanghai Jiao Tong University is supported by the National Key R\&D Program of China (2022ZD0114804), the National Natural Science Foundation of China (No. 62106141), and the Shanghai Sailing Program (21YF1421900). This work is done during Yuanfu’s internship at Shanghai Artificial Intelligence Laboratory. We sincerely thank our colleagues for engaging in insightful discussions and providing valuable feedback, with a particular acknowledgement to Ming Zhou and Zhanhui Zhou

\bibliography{aaai24}

\clearpage

\input{sections/appendix}

\end{document}

%% file: sections/0_abs.tex

\begin{abstract} \label{abstract}

Recent advancements in offline reinforcement learning (RL) have underscored the capabilities of Return-Conditioned Supervised Learning (RCSL), 
a paradigm that learns the action distribution based on target returns for each state in a supervised manner.
However, prevailing RCSL methods largely focus on deterministic trajectory modeling, disregarding stochastic state transitions and the diversity of future trajectory distributions.
A fundamental challenge arises from the inconsistency between the sampled returns within individual trajectories and the expected returns across multiple trajectories.
Fortunately, value-based methods offer a solution by leveraging a value function to approximate the expected returns, thereby addressing the inconsistency effectively.
Building upon these insights, we propose a novel approach, termed the Critic-Guided Decision Transformer (CGDT), which combines the predictability of long-term returns from value-based methods with the trajectory modeling capability of the Decision Transformer.
By incorporating a learned value function, known as the critic, CGDT ensures a direct alignment between the specified target returns and the expected returns of actions. This integration bridges the gap between the deterministic nature of RCSL and the probabilistic characteristics of value-based methods.
Empirical evaluations on stochastic environments and D4RL benchmark datasets demonstrate the superiority of CGDT over traditional RCSL methods.
These results highlight the potential of CGDT to advance the state of the art in offline RL and extend the applicability of RCSL to a wide range of RL tasks.
\end{abstract}

%% file: sections/1_intro.tex
\begin{figure}[t]
\centering
\includegraphics[width=0.99\columnwidth]{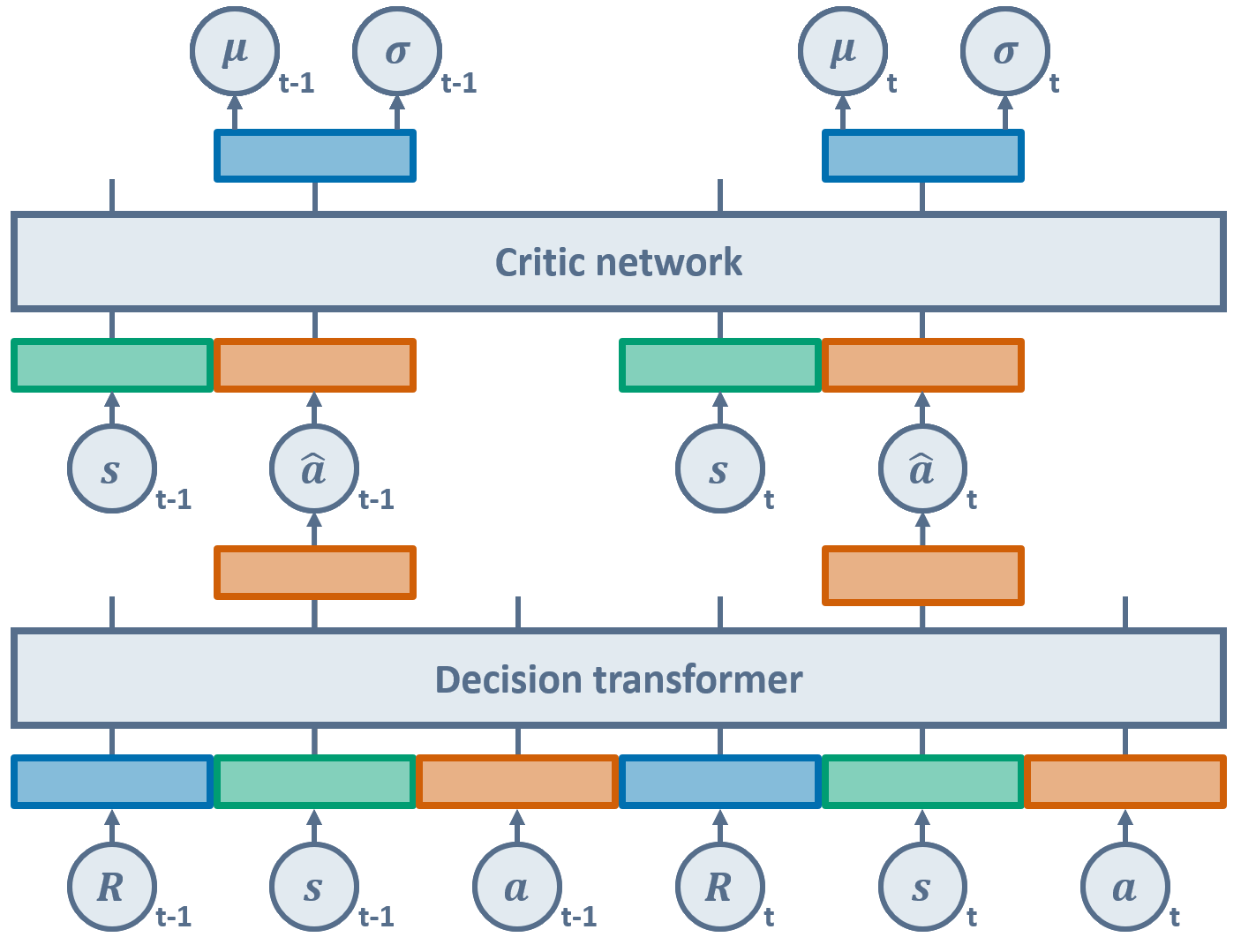}
\caption{\textit{Critic-Guided Decision Transformer framework}. The lower part is a vanilla Decision Transformer that takes the states $s$, actions $a$, and target returns $R$ as inputs to predict the next action $\hat{a}_t$ for each state $s_t$. The predicted actions are then passed through a critic, which is a Gaussian distribution with expected return mean $\mu_t$ and variance $\sigma_t$ learned from offline data. By minimizing the distance between the expected returns of the predicted actions and the target returns, e.g. ${\|( R_t - \mu_t ) / \sigma_t \|}_2$, the critic guides the policy to take actions that are consistent with the target returns.}
\label{fig2}
\end{figure}

\section{Introduction} \label{introduction}

Offline reinforcement learning (RL) addresses the problem of deriving effective policies from existing datasets that capture agent behaviors without interactions with environments.
A naive solution to offline RL is imitation learning (IL) \citep{hussein2017imitation}, which aims to emulate the behaviors of policies represented in the dataset. However, IL is limited in its ability to distinguish between optimal and suboptimal trajectories without predefined returns, often resulting in suboptimal policies that mirror the distribution of the training data.
To overcome the limitations of IL, recent research has introduced Return-Conditioned Supervised Learning (RCSL).
Representative works such as the Decision Transformer (DT) \citep{chen2021decision} and RvS \citep{emmons2021rvs} take cumulative returns or average returns as conditions to train a conditional policy that can differentiate desired (optimal) behaviors from the dataset.

However, RCSL struggles in stochastic environments and scenarios that require stitching abilities \citep{paster2022you}, where the policy needs to combine actions from suboptimal trajectories. 
The core issue limiting the performance of RCSL is the inconsistency between the sampled target (desired) returns and the expected returns of actions. 
In other words, trajectories with higher returns does not necessarily imply that their actions are superior to others; they can be a result of luck.
RCSL treats the return-to-go (RTG) as a quantity tied to a single trajectory, neglecting the stochastic state transitions and the broader distribution of future outcomes \citep{brandfonbrener2022does, bhargava2023sequence}.
This inconsistency is further exacerbated by the inherent uncertainty and approximation errors within behavior policies, resulting in inferior performance in stitching problems where suboptimal data present.

Fortunately, value-based methods \citep{sutton1998introduction}, on the other hand, provide a robust solution to handle this inconsistency. These methods estimate the expected cumulative returns of actions for each state, enabling an agent to choose optimal actions that maximize long-term returns. Q-learning algorithms \citep{kumar2020conservative, peng2019advantage, kostrikov2021offline, xu2023offline}, in particular, utilize temporal difference (TD) updates to learn a value function, allowing effective policy learning even in stochastic environments with highly suboptimal trajectories.

To address the limitations of RCSL, we propose a novel approach called the Critic-Guided Decision Transformer (CGDT). Our approach combines the predictability of long-term returns from value-based methods with the Decision Transformer framework. By utilizing a value function, known as the critic, to guide policy training, CGDT ensures that the expected returns of actions align with the specified target returns. This integration enables CGDT to effectively handle both stochastic environments and stitching scenarios, while still allowing conditional action selection. 

In this paper, we evaluate our proposed approach on stochastic environments and D4RL benchmark datasets. Our experimental results demonstrate significant improvements over pure RCSL in both stochastic environments and stitching problems. This showcases the potential of our method to advance the state-of-the-art in offline RL. Furthermore, our proposed approach holds promise for various RL tasks. We summarize our contributions as follows:
\begin{itemize}
    \item We provide an intuitive explanation for the pitfalls of RCSL in stochastic environments and stitching scenarios, arising from the inconsistency between the target (desired) returns and the expected returns of actions.
    
    \item We propose a novel approach, Critic-Guided Decision Transformer (CGDT), which leverages a critic to handle stochasticity from environments and uncertainty from suboptimal data while preserving the capability to act on variable conditional inputs.
    
    \item We evaluate our method on various benchmarks, including a Bernoulli Bandit game with stochastic rewards and D4RL benchmark datasets, and analyze how the use of the critic handles stochasticity and benefits CGDT.
\end{itemize}

%% file: sections/2_related.tex
\section{Related Work} \label{related}

\subsection{Offline RL}

Offline RL has seen the emergence of several methodologies to address the challenges of learning from fixed datasets \citep{prudencio2023survey}. These methodologies can be categorized into value-free and value-based approaches.
Value-free approaches do not necessarily rely on value functions. One such approach is \textit{Imitation learning} \citep{hussein2017imitation}, which aims to imitate the behavior policy by training on collected or desired trajectories filtered by heuristics or value functions \citep{chen2020bail, wang2020critic}.
\textit{Trajectory Optimization}, e.g. Multi-Game Decision Transformer (MGDT) \citep{lee2022multi} and Trajectory Transformer (TT) \citep{janner2021offline}, models joint state-action distribution over complete trajectories, reducing out-of-distribution (OOD) action selection. To enable effective planning, this approach utilizes techniques such as beam search and reward estimates.
Contrarily, value-based methods rely on value functions. Two common approaches in this category are \textit{Policy Constraints} and \textit{Regularization}. Policy Constraints ensure that the learned policy remains close to the behavioral policy, either through direct estimation \citep{fujimoto2019off, kostrikov2021fisher} or implicit modifications of the learning objective \citep{kumar2019stabilizing, peng2019advantage}. 
Regularization, e.g. CQL \citep{kumar2020conservative} and IQL \citep{kostrikov2021offline}, introduces penalty terms to influence policy behaviors without explicitly estimating behavioral policy.

\subsection{Return-Conditioned Supervised Learning}

Return-Conditioned Supervised Learning (RCSL) is a newly emergent class of algorithms that learns action distribution based on future returns statistics for each state via supervised learning \citep{schmidhuber2019reinforcement, kumar2019reward}. By conditioning on target returns, the policy can generate actions that closely resemble the behaviors presented in the dataset.
Decision Transformers (DT) and its variants \citep{siebenborn2022crucial, zheng2022online, hu2023graph, wen2023large} use returns-to-go, i.e. cumulative future returns, as the conditional inputs and model trajectories with causal transformers \citep{vaswani2017attention}.
RvS \citep{emmons2021rvs} investigates the effectiveness of conditioning on future states and average rewards.
These approaches explore the capabilities of different conditional inputs in various environments.
Generalized Decision Transformer \citep{furuta2021generalized} reveals that all these conditional supervised learning approaches are doing \textit{hindsight information matching} (HIM) indeed, i.e. to match the output trajectories with future information statistics.



%% file: sections/3_problem.tex
\begin{figure*}[t]
\centering
\includegraphics[width=0.99\textwidth]{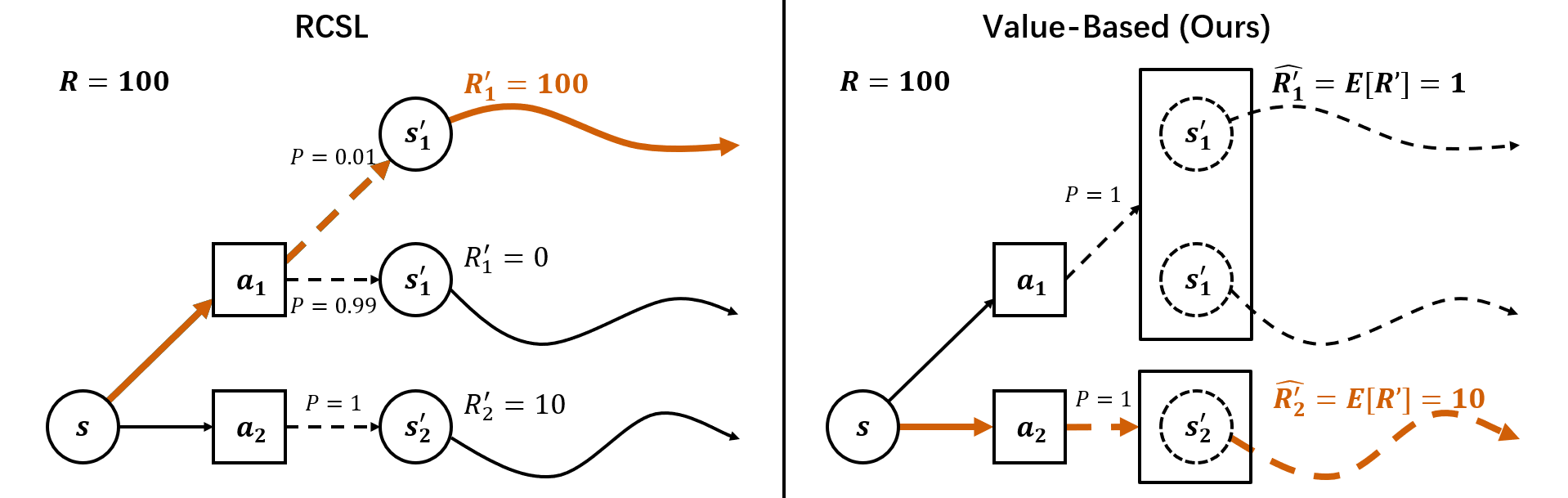}
\caption{\textit{Illustration of RCSL and Value-Based}. Considering a MDP where taking action $a_1$ has a low probability $P=0.01$ of leading to a future state $s'_1$ with high return $R'_1=100$, while the optimal action $a_2$ deterministically results in a future state $s'_2$ with return $R'_2=10$. \textit{Left}: RCSL selects action $a_1$ based on the target return $R=100$, regardless of its low probability, following the action distribution in the dataset. \textit{Right}: Value-based methods utilize a value function to estimate the expected returns over multiple trajectories and guide the policy to take actions aligned with the target return $R=100$. The colored arrows indicate the learned behaviors conditioning on the specified target return $R=100$, with both actions sampled in the dataset.}
\label{fig1}
\end{figure*}

\section{Pitfalls of RCSL}

In this section, we provide an intuitive explanation for the underlying reasons behind the limitations of RCSL in offline RL, particularly in the context of stochastic environments and scenarios involving stitching abilities. By focusing on these specific scenarios, we aim to reveal the fundamental factors that contribute to RCSL's failures in these settings.

\subsection{Stochasticity of Transitions}

RCSL faces significant challenges when applied to stochastic environments, even when provided with infinite data and without any approximation errors \citep{paster2022you, brandfonbrener2022does}. 
Consider the MDP depicted in Figure~\ref{fig1}, which presents two available actions, $a_1$ and $a_2$. 
Taking action $a_1$ leads to a future state $s'_1$ with a low probability $P=0.01$ of attaining high returns $R_1'=100$, while the remaining probability yields no returns. 
Conversely, taking action $a_2$ guarantees a future state $s'_2$ with returns $R_2'=10$, making it the optimal choice. 
RCSL (left of Figure~\ref{fig1}) models individual trajectories without considering transition probabilities. As a result, it may select the suboptimal action $a_1$ to attain the target return of $R=100$, despite its low probabilities, following the action distribution in the dataset.
This example illustrates that \textbf{trajectories with higher returns do not imply that their actions are superior to others; they could be a result of luck}.

To address the pitfall of RCSL in stochastic environments, alternative approaches have been proposed. 
For instance, \citet{paster2022you} propose environment-stochasticity-independent representations (ESPER), which cluster trajectories and utilize the average cluster returns as conditions for the policy.
Similarly, Q-Learning Decision Transformer (QDT) \citep{yamagata2023q}, utilizes a conservative value function to relabel return-to-go in the dataset.
DoC \citep{yang2022dichotomy} conditions the policy on a latent representation of future trajectories that is agnostic to stochasticity, achieved by minimizing mutual information. 
These approaches reply on \textit{Stochasticity-Independent Representations}, by defining or learning a representation of future trajectories that is independent of environment stochasticity.

Although these methods exhibit efficacy in stochastic contexts, they come with their limitations.
Methods that rely on relabeling returns-to-go (ESPER, QDT) may struggle with unmatched return-to-go values during inference. DoC requires complex objectives for representation learning and additional steps like sampling and value estimation..

\subsection{Stitching in Offline RL}

Indeed, the limitations of RCSL are not confined to stochastic settings. RCSL methods may exhibit suboptimal performance even in deterministic environments when suboptimal data is prevalent \citep{li2023survey}. 
In deterministic environments, \textbf{the uncertainty and approximation errors within the behavior policy introduce a form of stochasticity that resembles environmental stochasticity}. 
As a result, a tradeoff emerges between aggressive actions that may yield high returns but are hard to replicate, and conservative actions that consistently offer moderate returns. RCSL often prioritize the former, resulting in suboptimal performance.

Essentially, methods based on \textit{Stochasticity-Independent Representations} leverage probabilistic statistics from multiple trajectories to guide action selection.
Elastic Decision Transformer \citep{wu2023elastic} dynamically adjusts the context length of DT during inference, allowing it to "stitch" with an optimal trajectorywhile incorporating a value function without replacing the return-to-go.
MGDT and TT use either reward estimates or value functions to sample optimal actions during the planning stage.

From our perspective, incorporating probabilistic statistics from multiple trajectories offers a promising solution for stochasticity and suboptimal data.
As depicted in Figure~\ref{fig1} (right), these methods guide policy behaviors with learned expected returns from the entire distribution of future trajectories. These approaches ensure that the output actions align with the desired target returns in statistic, resolving the inconsistency arising from trajectory-level modeling.



%% file: sections/4_method.tex
\section{Method}
In this section, we first introduce the preliminary notations for offline RL and RCSL. Then, we propose the learning objective of training critics from the offline dataset and the learning objective of training policy with critic guidance which ensures the expected returns of actions are consistent with desired returns.
Finally, we introduce Critic-Guided Decision Transformer, a practical framework for optimizing policy with critic-guided learning objectives.

\subsection{Preliminaries}
In the offline RL setup, the objective is to train an agent solely from an existing dataset $\mathcal{D}$ of trajectories $\tau = (s_0, a_0, r_0, \cdots, s_{T-1}, a_{T-1}, r_{T-1})$ of states $s_t \in \mathcal{S}$, actions $a_t \in \mathcal{A}$, and rewards $r_t \in \mathcal{R}$ sampled by a behavior policy $\pi_\beta$ interacting with a finite horizon Markov Decision Process \citep{sutton1998introduction} with horizon $T$. We use $\tau_{i:j} := (s_t, a_t, r_t)_{t=i}^j$ to denote a sub-trajectory. Let $R(\tau) = \sum_{t=0}^T r_{t}$ denotes the cumulative return of the trajectory $\tau$. The goal of RL is to learn a policy that maximizes the expected cumulative return $\mathbb{E}\left[R(\tau)\right] = \mathbb{E}\left[\sum_{t=1}^T r_{t}\right]$.

In RCSL, we denote the return of a trajectory at timestep $t$ as
$R_t := \sum_{t'=t}^T r_{t'}$. $R_0$ is also known as the target (desired) return.
Let $\pi_\theta$ denotes the learning policy parameterized by $\theta$.
The objective of RCSL is typically to minimize the empirical negative log-likelihood loss (NLL), given by:
$$ \mathcal{L}(\theta) = \mathbb{E}_{\tau \in \mathcal{D}} \left[ - \sum_{0 \leq t < H} \log \left( \pi_\theta (a_t | \tau_{0:t-1}, s_t, R_t) \right) \right] .$$
During inference, the target return is substituted with a manually specified target return, often chosen as the maximum return among the trajectories in the dataset.

\subsection{Asymmetric Critic Training}
Following Bayes' rule, we can express the probability of taking an action $a_t$, given a return $R_t$ and a state $s_t$ as:
$p(a_t|R_t, s_t) \varpropto p(a_t|s_t)p(R_t|s_t, a_t)$, 
which suggests that to guide the action selection towards desired returns, we can model the probability distribution $p(R_t|s_t, a_t)$. Nonetheless, this distribution is typically unknown. Instead, we propose a parameterized critic $Q_\phi(R_t|\tau_{0:t-1}, s_t, a_t)$, which approximates $p(R_t|s_t, a_t)$ as a Gaussian distribution with learnable mean and variance \citep{bellemare2017distributional}. This critic is trained using an offline dataset $\mathcal{D}$ with NLL loss as the objective:
\begin{equation}
\label{nll}
    \mathcal{L}(\phi) = - \log Q_\phi(R_t|\tau_{0:t-1}, s_t, a_t).
\end{equation}

However, the quality of the data in $\mathcal{D}$ can be unbalanced, containing both optimal and suboptimal trajectories. To address this issue, we introduce an asymmetric NLL loss as the revised learning objective of fitting the critic:
\begin{equation}
\label{return}
\scalebox{0.99}{
    $ \mathcal{L}_{Q}(\phi) = - |\tau_c - \mathbb{I}( u > 0 )| \log Q_\phi(R_t|\tau_{0:t-1}, s_t, a_t),$
}
\end{equation}
where $ u = (R_t - \mu_t) / \sigma_t $, and $ \left( \mu_t, \sigma_t \right) \sim Q_\phi(\cdot |\tau_{0:t-1}, s_t, a_t)$ represent the mean and variance of the estimated return at current state $s_t$ and action $a_t$, respectively. Here, $\tau_c \in (0, 1)$ is an adjustable coefficient that controls the asymmetry of the loss. When $\tau_c > 0.5$, the critic is biased towards fitting optimal trajectories., while $\tau_c < 0.5$ biases the critic towards suboptimal trajectories. Setting $\tau_c = 0.5$ corresponds to using a scaled standard NLL loss in Equation \ref{nll}.

\subsection{Asymmetric Critic Guidance}
To encourage the selection of optimistic actions with expected returns higher than the target returns, we adopt the approach of \textit{Expectile Regression}, a variant of mean regression commonly used for estimating statistics of a random variable. Inspired by IQL \citep{kostrikov2021offline}, our method utilizes the critic to guide action selection using the following objective:
\begin{equation}
\label{expectile}
    \mathcal{L}_2^{\tau_p} (u) = |\tau_p - \mathbb{I}(u < 0)| u^2,
\end{equation}
where $u = (R_t - \mu_t) / \sigma_t $ and $ (\mu_t, \sigma_t) \sim Q_\phi(\cdot |\tau_{0:t-1}, s_t, \hat{a}_t)$. Here, $\hat{a}_t$ is sampled from the policy $\pi_\theta(\cdot | \tau_{0:t-1}, s_t, R_t)$. The variables $\mu_t$ and $\sigma_t$ represent the mean and variance of the estimated return at the current state $s_t$ and predicted action $\hat{a}_t$, respectively. The adjustable coefficient $\tau_p$ lies in the range $(0, 1)$.
When $\tau_p = 0.5$, it is equivalent to mean regression, which estimates the mean of the random variables. By adjusting $\tau_p$, we introduce asymmetry into the mean regression. In Equation~\ref{expectile}, a large $\tau_p > 0.5$ approximates a lower expectile of the advantage of estimated expected returns over target returns, i.e. $u = (R_t - \mu_t) / \sigma_t > 0$, vice versa. Consequently, it guides the policy to select optimistic actions with higher expected returns than those conditioned on.

\begin{algorithm}[tb]
\caption{Critic-Guided Decision Transformer}
\label{alg:algorithm1}
\textbf{Input}: Offline dataset $\mathcal{D}$, critic $Q_\phi$, policy $\pi_\theta$, iterations $M$, $N$, asymmetric critic coefficient $\tau_c$, expectile regression parameter $\tau_p$, and balance weight $\alpha$.
\begin{algorithmic} 
\STATE $\backslash\backslash$ Asymmetric Critic Training
\FOR{$i=1, ..., M$}
\STATE Sample a batch of trajectories $(s_t, a_t, r_t)$ from $\mathcal{D}$;
\STATE Compute return of sub-trajectory $\tau_{t:T}$, $R_t = \sum_t^T r_t$;
\STATE Update $Q_\phi$ with gradient:
\begin{align}
\mathbb{E}_{(s_t, a_t, R_t)} \left[ \nabla_\phi \mathcal{L}_{Q}(\phi) \right]; \nonumber
\end{align}
\ENDFOR
\STATE $\backslash\backslash$ Critic-Guided Policy Training
\STATE $\alpha' \leftarrow 0$
\FOR{$j=1, ..., N$}
\STATE $\alpha' \leftarrow \alpha' + \alpha / N$
\STATE Sample a batch of trajectories $(s_t, a_t, r_t)$ from $\mathcal{D}$;
\STATE Compute return of sub-trajectory $\tau_{t:T}$, $ R_t = \sum_t^T r_t$;
\STATE Predict action $\hat{a}_t \sim \pi_\theta(\cdot | \tau_{0:t-1}, s_t, R_t)$;
\STATE Predict return $(\mu_t, \sigma_t) \sim Q_\phi(\cdot |\tau_{0:t-1}, s_t, \hat{a}_t)$;
\STATE Compute \textit{expectile regression} loss:
$\mathcal{L}_2^{\tau_p}(\frac{R_t - \mu_t}{\sigma_t})$;

\STATE Update $\pi_\theta$ with gradient:
\begin{align}
\mathbb{E}_{(s_t, a_t, \hat{a}_t, R_t)} \left[ \nabla_\theta \mathcal{L}_{2}(a_t, \hat{a}_t) + \alpha' \nabla_\theta \mathcal{L}_2^{\tau_p}(\frac{R_t - \mu_t}{\sigma_t}) \right]; \nonumber
\end{align}
\ENDFOR
\STATE \textbf{return} $\pi_\theta$
\end{algorithmic}
\end{algorithm}

\subsection{Critic-Guided Decision Transformer}
Building upon the proposed learning objectives for critic training and critic guidance, we present the Critic-Guided Decision Transformer framework in Figure~\ref{fig2}, which provides a practical approach for optimizing policy with critic-guided learning objective.
Firstly, we evaluate the behavior policy by training a critic that estimates the cumulative returns of actions in the dataset through revised maximum likelihood estimation, as described in Equation \ref{return}. Then, we proceed with one-step policy improvement by optimizing the policy to select actions with expected returns that are consistent with or slightly better than the target returns.
This approach eliminates the need for off-policy evaluation and has the potential to yield the same policy learned by multi-step critic regularization methods \citep{eysenbach2023connection}.





Notably, the critic trained on the offline dataset may suffer from overestimating out-of-distribution actions. Solely optimizing the critic guidance term (Equation~\ref{expectile}) may mislead the policy towards overestimated actions. To mitigate this problem and ensure that the policy remains close to the data distribution, we introduce the vanilla RCSL learning objective as a constraint. In practice, this objective is implemented as an $l^2$-norm term on actions, discouraging the selection of actions that deviate significantly from the data distribution. Consequently, the overall policy loss is formulated as:
\begin{equation}
\scalebox{1.0}{
    $ \mathcal{L}_{\pi}(\theta; \alpha) = \mathcal{L}_2(a_t, \hat{a}_t) + \alpha \cdot \mathcal{L}_2^{\tau_p} (\frac{R_t - \mu_t}{\sigma_t}) $,
}
\end{equation}
where $\hat{a}_t \sim \pi_\theta(\cdot | \tau_{0:t-1}, s_t, R_t)$, and ($\mu_t, \sigma_t$) are sampled from the critic $Q_\phi$. Additionally, $\alpha$ is a balance weight.

In practical implementation, we utilize validation errors as a means to detect overfitting during critic training, which can monitor the performance of the critic model and apply early stopping to mitigate overestimation issues.
Additionally, we introduce a linearly increasing weight coefficient to balance the action regression loss and the critic guidance term during policy training. This coefficient eliminates the need for manual tuning and provides a mechanism to control policy behavior automatically.
The algorithm implementation details are summarized in Algorithm~\ref{alg:algorithm1}.

%% file: sections/5_exp.tex
\begin{table*}[t]
\centering
\begin{tabular}{ll|rrrrrrr|rr}
\hline
\multicolumn{1}{c}{\textbf{Dataset}} & 
\multicolumn{1}{c|}{\textbf{Environment}} & 
\multicolumn{1}{c}{\textbf{10\%BC$^*$}} & 
\multicolumn{1}{c}{\textbf{RvS$^*$}} & 
\multicolumn{1}{c}{\textbf{DT$^*$}} & 
\multicolumn{1}{c}{\textbf{QDT$^*$}} & 
\multicolumn{1}{c}{\textbf{CQL$^*$}} & 
\multicolumn{1}{c}{\textbf{IQL$^*$}} & 
\multicolumn{1}{c}{\textbf{TT$^*$}} & 
\multicolumn{1}{|c}{\textbf{DT}} & 
\multicolumn{1}{c}{\textbf{CGDT}} \\
\hline
Medium & Halfcheetah        &  42.5 &  41.6 &  42.6 & 42.2 &  44.0 &  \textbf{47.4} &  46.9 & 42.7 & \textbf{43.0} \\
Medium & Hopper             &  56.9 &  60.2 &  \textbf{67.6} & 65.3 &  58.5 &  66.3 &  61.1 & 67.5 & \textbf{96.9} \\
Medium & Walker2d           &  75.0 &  71.7 &  74.0 & 70.1 &  72.5 &  78.3 &  \textbf{79.0} & 76.8  & \textbf{79.1} \\
\hline
Medium-Replay & Halfcheetah &  40.6 &  38.0 &  36.6 & 35.7 & \textbf{45.5} & 44.2 &  41.9 & 40.2 & \textbf{40.4} \\
Medium-Replay & Hopper      &  75.9 &  73.5 &  82.7 & 55.3 & \textbf{95.0} & 94.7 &  91.5 & 88.3 & \textbf{93.4} \\
Medium-Replay & Walker2d    &  62.5 &  60.6 &  66.6 & 59.1 & 77.2 & 73.9 &  \textbf{82.6} & 73.0 & \textbf{78.1} \\
\hline
Medium-Expert & Halfcheetah &  92.9 &  92.2 &  86.8 &    / & 91.6 & 86.7 &  \textbf{95.0} & 93.1 & \textbf{93.6} \\
Medium-Expert & Hopper      & \textbf{110.9} & 101.7 & 107.6 &    / & 105.4 & 91.5 & 110.0 & \textbf{108.6} & 107.6 \\
Medium-Expert & Walker2d    & 109.0 & 106.0 & 108.1 &    / & 108.8 & \textbf{109.6} & 101.9 & 109.0 & \textbf{109.3} \\
\hline
\multicolumn{2}{c|}{\textbf{Sum}} & 666.2 & 645.5 & 672.6 & / & 698.5 & 692.6 & \textbf{722.5} & 699.2 & \textbf{741.5} \\
\hline
Umaze & Antmaze             &  62.8 & 64.4 &  59.2 &    / & 74.0 & 87.5 & \textbf{100.0} & 61.0 & \textbf{71.0} \\
Umaze-Diverse & Antmaze     &  50.2 & 70.1 &  53.0 &    / & \textbf{84.0} & 62.2 &     / & 55.0 & \textbf{71.0} \\
\hline
\multicolumn{2}{c|}{\textbf{Sum}} & 113.0 & 134.5 & 112.2 & / & \textbf{158.0} & 149.7 & / & 116.0 & \textbf{142.0} \\
\hline
\end{tabular}
\caption{\textit{Overall performance}. The Critic-Guided Decision Transformer (CGDT) demonstrates competitive or superior performance compared to prior offline RL algorithms on D4RL datasets. 
Particularly, on \textit{medium} and \textit{medium-replay} datasets where suboptimal data present, CGDT significantly outperforms RCSL methods such as RvS, DT, and QDT.
Its performance is on par with value-based algorithms such as CQL and IQL, and trajectory optimization algorithms such as TT. Average normalized scores over 5 random seeds are reported, each evaluated for 100 episodes; $^*$baseline results are taken from original papers.}
\label{table1}
\end{table*}

\begin{table*}[t]
\centering
\begin{tabular}{ll|rrr|rrr}
\hline
\multicolumn{1}{c}{\textbf{Dataset}} & 
\multicolumn{1}{c|}{\textbf{Environment}} & 
\multicolumn{1}{c}{\textbf{DT (sparse)}} & 
\multicolumn{1}{c}{\textbf{CGDT (sparse)}} & 
\multicolumn{1}{c|}{\textbf{$\mathbf{\delta_{sparse}}$}} & 
\multicolumn{1}{c}{\textbf{DT (dense)}} & 
\multicolumn{1}{c}{\textbf{CGDT (dense)}} & 
\multicolumn{1}{c}{\textbf{$\mathbf{\delta_{dense}}$}}  \\
\hline
Medium & Halfcheetah        &  42.7 &  \textbf{43.1} &  0.4 &  42.7 &  \textbf{43.0} &  0.3 \\
Medium & Hopper             &  65.9 &  \textbf{78.1} & 12.2 &  67.5 &  \textbf{96.9} & 29.4 \\
Medium & Walker2d           &  76.9 &  \textbf{79.9} &  3.0 &  76.8 &  \textbf{79.1} &  2.3 \\
\hline
Medium-Replay & Halfcheetah &  \textbf{40.9} &  40.2 & -0.7 &  40.2 &  \textbf{40.4} &  0.2 \\
Medium-Replay & Hopper      &  84.5 &  \textbf{86.3} &  1.8 &  88.3 &  \textbf{93.4} &  5.1 \\
Medium-Replay & Walker2d    &  69.4 &  \textbf{73.9} &  4.5 &  73.0 &  \textbf{78.1} &  5.2 \\
\hline
Medium-Expert & Halfcheetah &  93.6 &  \textbf{93.6} &  0.0 &  93.1 &  \textbf{93.6} &  0.5 \\
Medium-Expert & Hopper      & \textbf{106.3} & 106.2 & -0.1 & \textbf{108.6} & 107.6 & -1.0 \\
Medium-Expert & Walker2d    & 107.3 & \textbf{109.4} &  2.1 & 109.0 & \textbf{109.3} &  0.4 \\
\hline
\multicolumn{2}{c|}{\textbf{Sum}} & 687.4 & \textbf{710.5} & 23.1 & 699.2 & \textbf{741.5} & 42.4 \\
\hline
\end{tabular}
\caption{\textit{Sparse Reward}. We evaluate the performance of Critic-Guided Decision Transformer (CGDT) in sparse (delayed) reward settings on D4RL locomotion tasks using the same hyperparameters as in dense reward settings, which might not be optimal.
Critic guidance benefits CGDT in sparse reward settings, while dense rewards lead to a larger improvement over sparse rewards.
Average normalized scores over 5 random seeds are reported, each evaluated for 100 episodes.}
\label{table2}
\end{table*}


\section{Experiments}
We conduct a series of experiments to address the following questions: \textbf{(i)} How effectively does CGDT handle stochasticity in the Bernoulli bandit problem with stochastic rewards? \textbf{(ii)} How does critic guidance benefit the performance of CGDT in the presence of suboptimal data with sparse and dense rewards? \textbf{(iii)} How consistent are the returns achieved by CGDT with the specified target returns? Additionally, we conduct ablation studies to examine the influence of asymmetry in critic training and critic guidance.

\subsection{Bernoulli Bandit (Stochasticity)}
\begin{figure}[t]
\centering
\includegraphics[width=1.0\columnwidth]{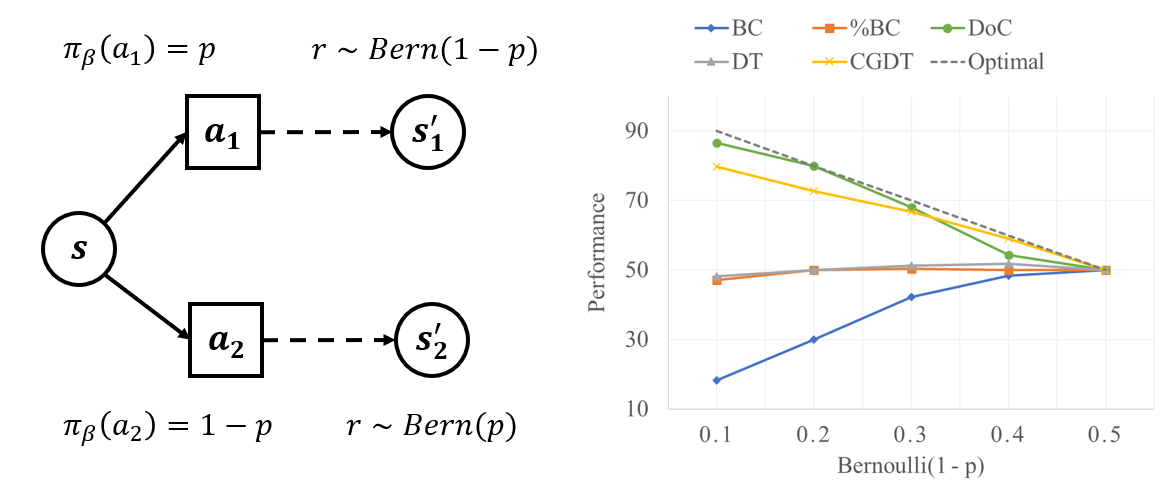} 
\caption{\textit{Bernoulli Bandit}. \textit{Left}: A two-armed bandit, following \citet{yang2022dichotomy}, to evaluate the performance of CGDT under environment with stochastic rewards. \textit{Right}: DoC, CGDT achieve close to Bayes-optimal (dotted), while BC and DT fail. Average normalized scores over 5 random seeds are reported, each evaluated for 1000 episodes.}
\label{fig3}
\end{figure}

To evaluate the ability of an agent to handle environmental stochasticity, we employ a two-armed Bernoulli bandit game with stochastic rewards, following \citet{yang2022dichotomy}. As illustrated in Figure \ref{fig3}, the game consists of two arms, denoted as $a_1$ and $a_2$, which generate stochastic rewards drawn from Bernoulli distributions of $Bern(1-p)$ and $Bern(p)$, respectively. Arm $a_1$ yields a non-zero reward with probability $1 - p$, while arm $a_2$ does so with probability $p$. In this setup, a smaller value of $p$ (i.e., $p < 0.5$) makes arm $a_1$ the optimal choice with a higher expected return. To ensure a balanced occurrence of successful retrievals for both arms in the offline dataset, the behavior policy pulls arm $a_1$ with probability $\pi_\beta(a_1 | s) = p$.

We implement our approach and baselines as stochastic policies, except for DoC, which samples actions from a prior distribution and uses a value function to distinguish the action with highest return. We train these methods using 10,000 samples, where the probability $p$ varied in the range $\{0.1, 0.2, 0.3, 0.4, 0.5\}$. In Figure \ref{fig3} (right), we observe that supervised learning (SL) methods (BC, \%BC) and RCSL (DT) methods converge to suboptimal behaviors without value functions. However, both CGDT and DoC achieve results close to the Bayes-optimal. The ability to handle stochasticity using a value function is a significant advantage of CGDT compared to SL and RCSL.

\subsection{OpenAI Gym (Stitching)}

\begin{figure*}[t]
\centering
\includegraphics[width=0.9\textwidth]{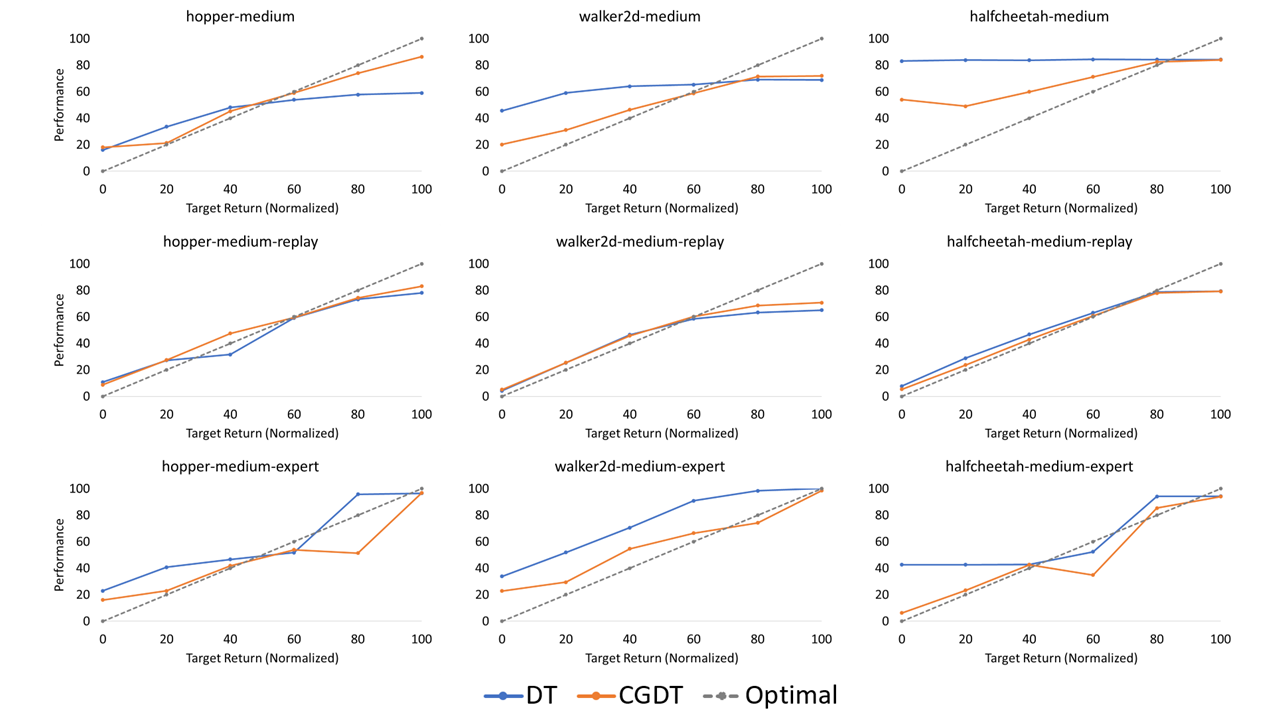} 
\caption{\textit{Conditional Behaviors}. Evaluation returns achieved by DT and CGDT when conditioned on the specified target returns. The dotted lines denote the optimal behaviors. It is observed that CGDT sticks to the target returns more closely than DT, which indicates a more consistent behavior observed in CGDT.}
\label{fig4}
\end{figure*}

To evaluate how critic guidance benefits CGDT in stitching problems, we conduct further experiments on the D4RL datasets \citep{fu2020d4rl}. These datasets provide standardized environments and various datasets with different qualities. For our evaluation, we specifically focus on the MuJoCo locomotion tasks using the \textit{medium}, \textit{medium-replay}, and \textit{medium-expert} datasets, along with navigation tasks. We compare our approach with SL and RCSL algorithms, e.g. \%BC, RvS, DT, and QDT. Besides, we also compare with value-based algorithms, e.g. CQL and IQL, and trajectory optimization algorithms like TT. 

We present the results in Table \ref{table1} with baseline results retrieved from the original papers. SL and RCSL methods perform well in high-quality datasets but struggle to achieve optimal performance in suboptimal datasets. Conversely, value-based methods (CQL and IQL) exhibit strong stitching abilities in suboptimal datasets but do not generalize well to high-quality datasets. Trajectory optimization, e.g. TT, demonstrates exceptional overall performance, showcasing the advantage of trajectory modeling. Our approach, CGDT, shows significant improvement over SL and RCSL methods in suboptimal data regimes, while maintaining its optimal performance in high-quality datasets.

To further investigate the advantages of utilizing a value function with limited reward signals, we evaluate our approach in scenarios with sparse (delayed) rewards, where the rewards are only granted at the final timestep of each trajectory. The results for the sparse reward case are presented in Table \ref{table2}. Notably, we do not extensively tune the hyperparameters and use the same settings as in the dense reward settings, which may not be optimal for the sparse reward scenario. Nonetheless, we observe improved performance over DT on the sparse reward tasks, though, with lesser improvement as in the dense reward case.  These findings indicate that critics trained in both sparse and dense reward settings contribute to the benefits of CGDT, with more substantial improvements observed in dense reward scenarios.


\subsection{Conditional Behaviors (Consistency)}
One significant difference between RCSL and value-based algorithms is the ability to behave conditionally. 
Conditional behaviors, which are characteristic of RCSL, are not typically exhibited by most value-based algorithms. This is because the limited model capacity and potentially conflicting learning objectives in training may jeopardize the optimal performance. However, the ability to exhibit conditional behaviors is essential for controllability and flexibility, enabling application in a wider range of scenarios.

We evaluate the conditional behaviors of our approach and DT under different target returns $R' = \lambda R$, where $\lambda \in \{ 0, 0.2, \cdots, 1 \} $ and $R$ represents the original target return.
From Figure \ref{fig4}, we observe that DT naturally exhibits conditional behaviors but shows weakened consistency, particularly when the datasets do not include suboptimal data (e.g., \textit{medium} and \textit{medium-expert} datasets). In contrast, CGDT sticks to target returns more closely, even in datasets where suboptimal data are absent. This result indicates an improved consistency of CGDT between the expected returns of actions and the target returns.

\subsection{Ablation Study}
To investigate the effect of asymmetry in critic training and policy training, we conduct a series of ablation experiments. Initially, we set the hyperparameters $\tau_c$ and $\tau_p$ to $0.5$. By varying $\tau_c$ and $\tau_p$ within the range of $[0.3, 0.7]$, we control the asymmetries during critic training and policy training, respectively. Notably, datasets with different distributions exhibit varying preferences. In general, during critic training, there is a tendency to fit the model towards high-quality data. On the other hand, during policy training, optimistic actions with higher expected returns are favored. The \textit{Antmaze} datasets present a special case, which primarily consist of truncated trajectories with zero returns, where low returns do not necessarily indicate suboptimal performance.

\begin{figure}[t]
\centering
\includegraphics[width=0.95\columnwidth]{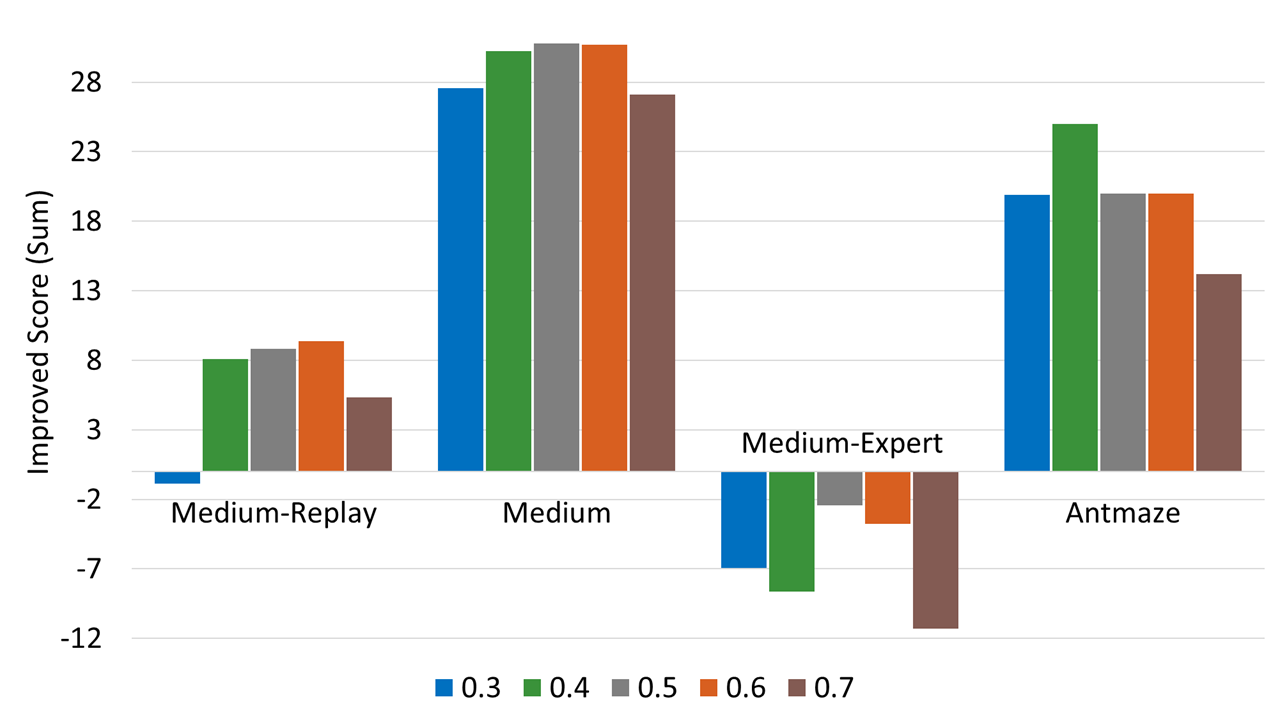}
\caption{\textit{Ablations of $\tau_c$ in critic training}. We show the improved performance of CGDT over DT with different $\tau_c \in \{ 0.3, 0.4, 0.5, 0.6, 0.7 \}$ in Equation~\ref{return}. A large $\tau_c$ indicates a bias over high-quality data, and vice versa.
}
\label{fig5}
\end{figure}

\begin{figure}[t]
\centering
\includegraphics[width=0.95\columnwidth]{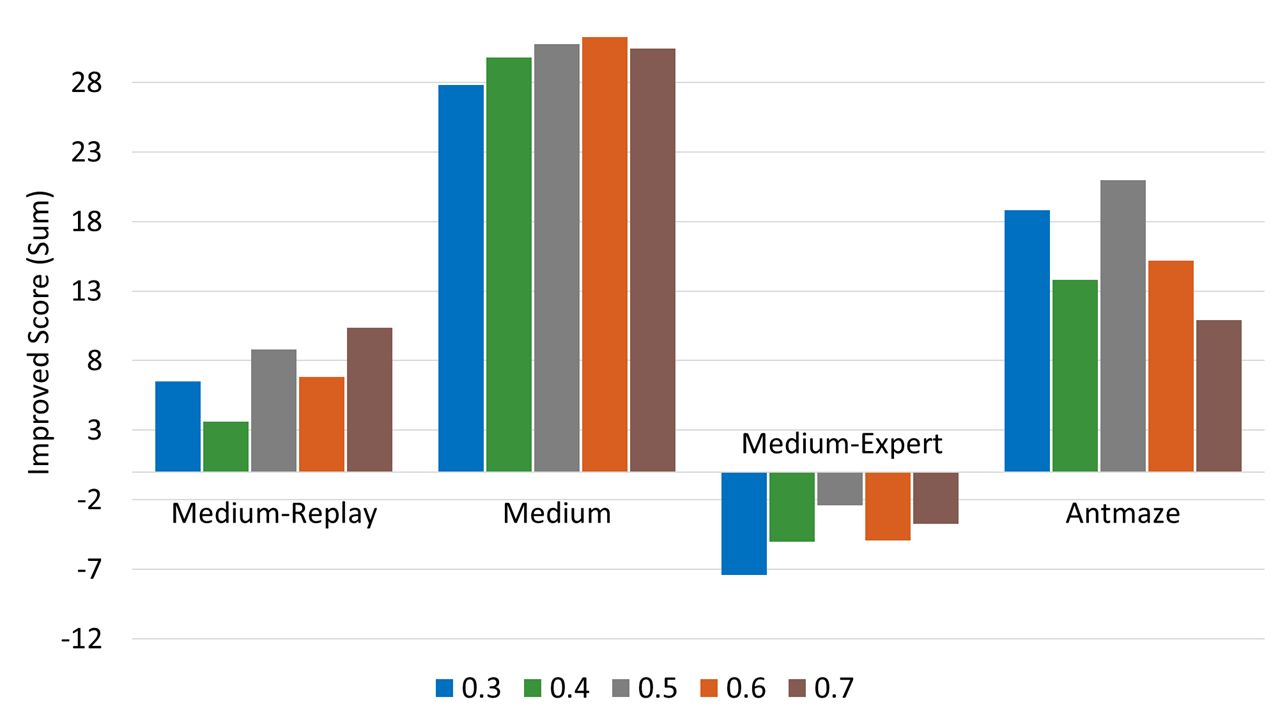}
\caption{\textit{Ablations of $\tau_p$ in policy training}. We show the improved performance of CGDT over DT with different $\tau_p \in \{ 0.3, 0.4, 0.5, 0.6, 0.7 \}$ in Equation~\ref{expectile}. A large $\tau_p$ favors actions with higher expected returns, and vice versa. 
}
\label{fig6}
\end{figure}

%% file: sections/6_conc.tex
\section{Conclusion}

We propose Critic-Guided Decision Transformer, a general framework for RCSL that utilizes a value function, to guide the trajectory modeling process. This framework effectively solves the inconsistency between the expected returns of actions and the target returns while preserving the conditional characteristic of RCSL. Our empirical results demonstrate that CGDT is highly capable of handling stochastic environments and addressing challenges in stitching problems posed by suboptimal data, which might provide fresh insights for extending the application of RCSL to broader domains.







%% file: sections/appendix.tex
\appendix

\section{Appendix}
In the appendix, we present more details on the implementation, results, and further analysis.

\subsection{Hyperparameters}
\label{sub:hyper}

We present the common hyperparameters utilized in both the DT and CGDT algorithms in Table~\ref{sup:table1} and \ref{sup:table2}. Our implementation of CGDT is based on the public codebase of the Online Decision Transformer (ODT) \citep{zheng2022online}.

In CGDT, both the critic and the policy are implemented as transformers \citep{vaswani2017attention}. We train the critic using 90\% of the trajectories from the dataset, reserving the remaining 10\% for validation purposes to detect overfitting. Early stopping is applied during the critic training process. To simplify the manual tuning of the balance weight, we introduce a linearly increasing balance weight during policy training.

All algorithms are trained with the LAMB optimizer \citep{You2019LargeBO} following the implementation of ODT, for which we also report the learning rates and weight decay parameters as follows. We basically use the same hyperparameters for the critic and the policy for simplicity.

\begin{table*}[htbp]
\centering
\begin{tabular}{ll}
\hline
\multicolumn{1}{l}{\textbf{Hyperparameter}} & 
\multicolumn{1}{l}{\textbf{Value}} \\
\hline
Number of layers                & 3 \\
Number of attention heads       & 4 \\
Embedding dimension             & 128 \\
Activation function             & Relu \\
Positional encoding             & Yes \\
Batch size                      & 256 \\
Context length $K$              & 1 for Bernoulli bandit \\
                                & 5 for Halfcheetah-Medium \\
                                & 10 for Halcheetah-\{Medium-Replay, Medium-Expert\}, Antmaze-Umaze \\
                                & 20 for all the other tasks \\
Dropout                         & 0.1 \\
Learning rate                   & 1e-3 for Halfcheetah \\
                                & 1e-4 for all the other tasks \\
Grad norm clip                  & 0.25 \\
Weight decay                    & 5e-4 \\
Learning rate warmup            & linear warmup for 10000 training steps \\
Eval episodes                   & 1000 for Bernoulli bandit \\
                                & 100 for all the other tasks \\
\hline
\end{tabular}

\caption{Common hyperparameters of DT and CGDT policy training and evaluation.}
\label{sup:table1}
\end{table*}

\begin{table*}[htbp]
\centering
\begin{tabular}{ll}
\hline
\multicolumn{1}{l}{\textbf{Hyperparameter}} & 
\multicolumn{1}{l}{\textbf{Value}} \\
\hline
Number of layers                & 2 \\
Number of attention heads       & 4 \\
Embedding dimension             & 128 \\
Activation function             & Relu \\
Positional encoding             & Yes \\
Batch size                      & 256 \\
Context length $K$              & 1 for Bernoulli bandit \\
                                & 5 for Halfcheetah-Medium \\
                                & 10 for Halcheetah-\{Medium-Replay, Medium-Expert\}, Antmaze-Umaze \\
                                & 20 for all the other tasks \\
Dropout                         & 0.1 \\
Learning rate                   & 1e-3 for Halfcheetah \\
                                & 1e-4 for all the other tasks \\
Grad norm clip                  & 0.25 \\
Weight decay                    & 5e-4 \\
Learning rate warmup            & linear warmup for 10000 training steps \\
\hline
\end{tabular}

\caption{Common hyperparameters of CGDT critic training. We basically use the same hyperparameters as in policy training for simplicity.}
\label{sup:table2}
\end{table*}

For all the experiments conducted on DT, we perform a final evaluation of each algorithm on various settings. The evaluation consists of running 100 episodes in the environment, using 5 random seeds for each evaluation. During this final evaluation, we explore different context lengths selected from the set $\{ 1, 5, 10, 15, 20 \}$, or returns-to-go (RTG) in different scales $\lambda \in \{ 0, 0.2, 0.4, \cdots, 2.0 \}$. The optimal evaluation settings for DT are summarized in Table~\ref{sup:table3}.

Regarding CGDT, the most crucial hyperparameters to tune are the tradeoff between the original RCSL learning objective and the critic guidance learning objective. The critic guidance term is influenced by parameter $\tau_c$, which controls the bias on data quality during critic training, and parameter $\tau_p$, which controls the preference for optimistic actions with higher expected returns in policy training. In addition to the aforementioned settings varied during the final evaluation of DT, we further tune different values of $\tau_c$ in the set $\{ 0.3, 0.4, 0.5, 0.6, 0.7 \}$ for critic training, or different values of $\tau_p$ in the set $\{ 0.1, 0.3, 0.4, 0.5, 0.6, 0.7, 0.9 \}$ for policy training. It's important to note that not all combinations of $\tau_c$ and $\tau_p$ are tuned. To simplify the tuning process and avoid meticulous manual adjustments, we also incorporate a linearly increasing weight coefficient that eliminates the need for careful tuning of the maximal balance weight $\alpha$. The optimal settings for CGDT are summarized in Table~\ref{sup:table4}.

\begin{table*}[htbp]
\begin{center}
\begin{tabular}{llccc}
\hline
\multicolumn{1}{l}{\textbf{Environment}} & 
\multicolumn{1}{l}{\textbf{Dataset}} & 
\multicolumn{1}{c}{\textbf{context length}} &
\multicolumn{1}{c}{\textbf{eval context length}} &
\multicolumn{1}{c}{\textbf{RTG}} \\
\hline
Hopper      & Medium        &  20 & 20 &  6480 \\
Hopper      & Medium-Replay &  20 & 20 &  4320 \\
Hopper      & Medium-Expert &  20 & 10 &  3600 \\
Walker2d    & Medium        &  20 & 20 & 10000 \\
Walker2d    & Medium-Replay &  20 & 20 & 10000 \\
Walker2d    & Medium-Expert &  20 & 20 &  7000 \\
Halfcheetah & Medium        &   5 &  5 &  8400 \\
Halfcheetah & Medium-Replay &  10 & 10 &  6000 \\
Halfcheetah & Medium-Expert &  10 & 10 &  6000 \\
Antmaze     & Umaze         &   5 &  5 &     1 \\
Antmaze     & Umaze-Diverse &  20 & 20 &     2 \\
\hline
\end{tabular}
\caption{Optimal hyperparameters for DT in each domain. The evaluation was conducted separately for various context lengths or Return-to-Go (RTG) settings during the final evaluation. The performance is measured using the average normalized scores over 5 seeds, with each seed evaluated for 100 episodes.}
\label{sup:table3}
\end{center}
\end{table*}

\begin{table*}[htbp]
\centering
\begin{tabular}{llcccccc}
\hline
\multicolumn{1}{c}{\textbf{Dataset}} & 
\multicolumn{1}{c}{\textbf{Environment}} & 
\multicolumn{1}{c}{\textbf{context length}} &
\multicolumn{1}{c}{\textbf{eval context length}} &
\multicolumn{1}{c}{\textbf{RTG}} &
\multicolumn{1}{c}{\textbf{$\tau_c$}} &
\multicolumn{1}{c}{\textbf{$\tau_p$}}  &
\multicolumn{1}{c}{\textbf{$\alpha$}} \\
\hline
Hopper      & Medium        &  20 & 15 &  3600 & 0.5 & 0.9 & 0.2  \\
Hopper      & Medium-Replay &  20 & 20 &  3600 & 0.5 & 0.7 & 0.12 \\
Hopper      & Medium-Expert &  20 & 10 &  3600 & 0.5 & 0.5 & 0.3  \\
Walker2d    & Medium        &  20 & 20 &  6000 & 0.5 & 0.6 & 0.4  \\
Walker2d    & Medium-Replay &  20 & 20 & 10000 & 0.5 & 0.7 & 0.06 \\
Walker2d    & Medium-Expert &  20 & 20 &  8000 & 0.5 & 0.9 & 0.3  \\
Halfcheetah & Medium        &   5 &  5 &  8400 & 0.5 & 0.3 & 0.2  \\
Halfcheetah & Medium-Replay &  10 & 10 &  6000 & 0.5 & 0.1 & 0.1  \\
Halfcheetah & Medium-Expert &  10 & 10 & 12000 & 0.6 & 0.5 & 0.2  \\
Antmaze     & Umaze         &  10 & 10 &   1.2 & 0.4 & 0.5 & 0.06 \\
Antmaze     & Umaze-Diverse &  20 &  5 &     1 & 0.5 & 0.8 & 0.02 \\
\hline
\end{tabular}
\caption{Optimal hyperparameters for CGDT in each domain. We additionally tune different values of $\tau_c$ in the set $\{ 0.3, 0.4, 0.5, 0.6, 0.7 \}$ for critic training, or different values of $\tau_p$ in the set $\{ 0.1, 0.3, 0.4, 0.5, 0.6, 0.7, 0.9 \}$ for policy training individually. The performance is measured using the average normalized scores over 5 seeds, with each seed evaluated for 100 episodes.}
\label{sup:table4}
\end{table*}

\subsection{Training Datasets}
The training dataset for the Bernoulli bandit game is generated using a behavior policy. This policy has probabilities $p \in \{0.1, 0.2, 0.3, 0.4\}$ for choosing arm $a_1$. When arm $a_1$ is selected, it has probabilities $p \in \{0.9, 0.8, 0.7, 0.6\}$ of yielding a reward of $1$, otherwise, it does not provide any reward. On the other hand, arm $a_2$ has probabilities $p \in \{0.1, 0.2, 0.3, 0.4\}$ to produce a reward of $1$. The arms are pulled a total of $10000$ times to collect the training dataset for each case.

\subsection{Sampling Strategy}
The sampling strategy utilized in CGDT is identical to the one employed in ODT. It involves initially sampling a single trajectory $\tau$ from the offline dataset $\mathcal{D}$, with the probability of selection being proportional to the length of the episode: $\mathcal{P}(\tau) \propto |\tau|$. Subsequently, subsequences are uniformly sampled from each selected trajectory.

\subsection{Training Curves}
\begin{figure*}[htbp]
\centering
\begin{subfigure}[b]{0.49\textwidth}
\includegraphics[width=\textwidth]{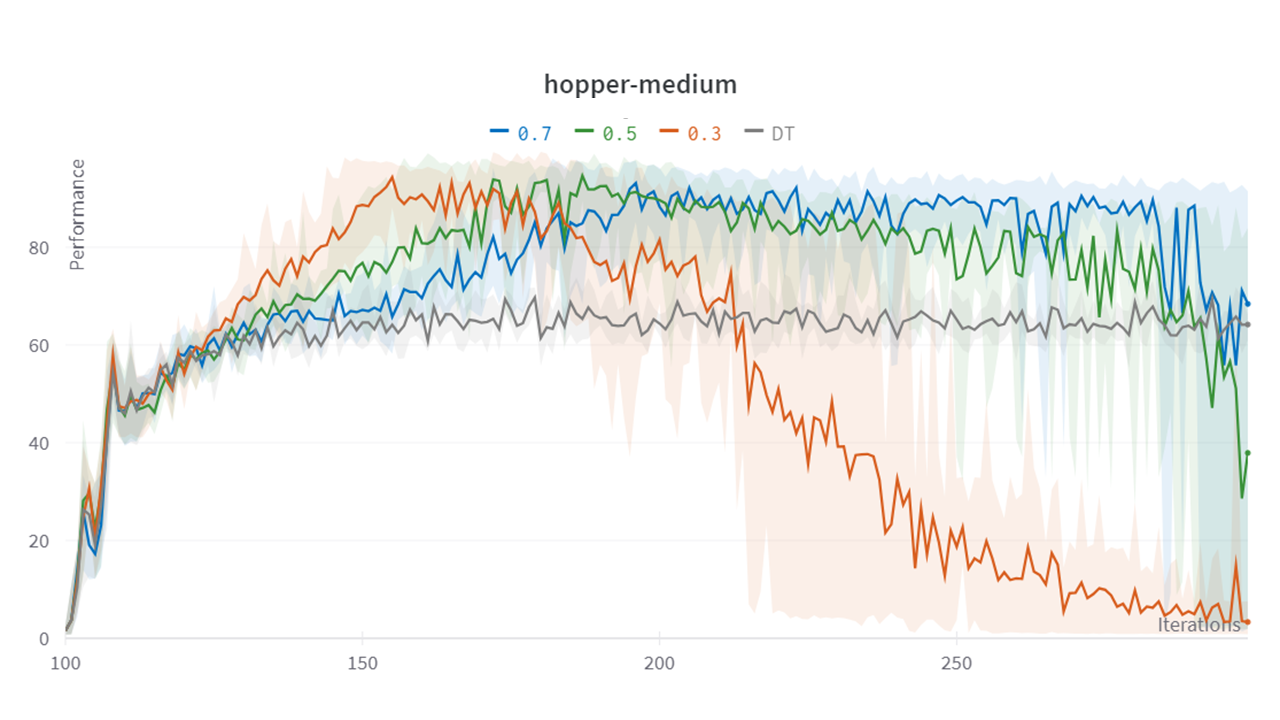}
\caption{$\tau_c \in \{0.3, 0.5, 0.7\}, \tau_p = 0.5$}
\end{subfigure}
\begin{subfigure}[b]{0.49\textwidth}
\includegraphics[width=\textwidth]{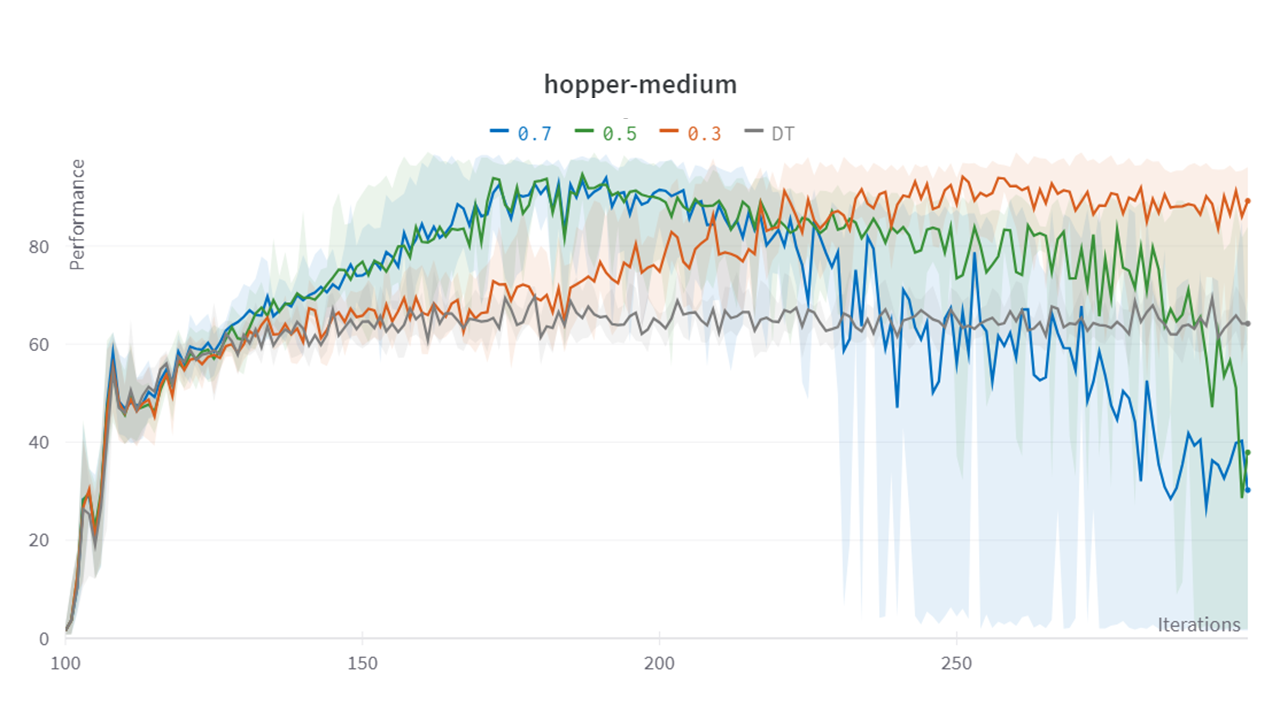}
\caption{$\tau_c = 0.5, \tau_p \in \{0.3, 0.5, 0.7\}$}
\end{subfigure}
\caption{Policy training curves with linearly increasing weight coefficient $\alpha'$ from $0$ to $\alpha$ in Hopper-Medium. The curves depict the performance of the policy during training with different values of $\tau_c$ and $\tau_p$ in the critic and policy training, respectively.}
\label{sup:fig_curve}
\end{figure*}

For different settings of $\tau_c$ and $\tau_p$, which control the asymmetry in critic and policy training, the optimal choice of the balance weight $\alpha$ can vary significantly..
To address this, we introduce a linearly increasing weight coefficient $\alpha'$ as a substitute for manual tuning of $\alpha$.
Figure~\ref{sup:fig_curve} illustrates the training curves for DT and CGDT on the \textit{Hopper} environment using the \textit{Medium} dataset under different settings.

We observe that the performance curves of CGDT policies are typically convex. Starting from the original RCSL learning objective, CGDT policies improve as the balance weights remain below certain thresholds. However, if the balance weights become too large, the critic guidance term dominates the policy training, resulting in suboptimal performance without the constraints of RCSL learning objectives. This suboptimal performance may be primarily attributed to the overestimation of out-of-distribution actions by the critic.

\subsection{Detailed Ablation Study}
\begin{figure*}[htbp]
\centering
\begin{subfigure}[b]{0.45\textwidth}
\includegraphics[width=\textwidth]{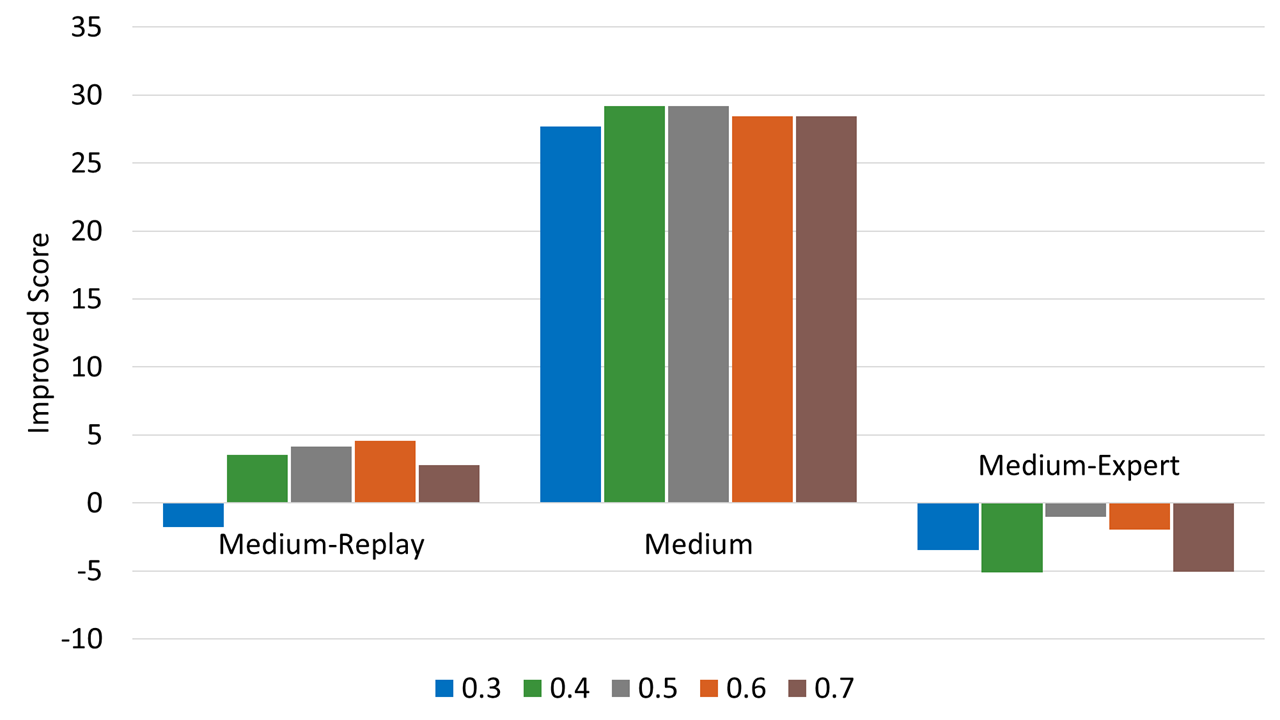}
\caption{Hopper}
\end{subfigure}
\begin{subfigure}[b]{0.45\textwidth}
\includegraphics[width=\textwidth]{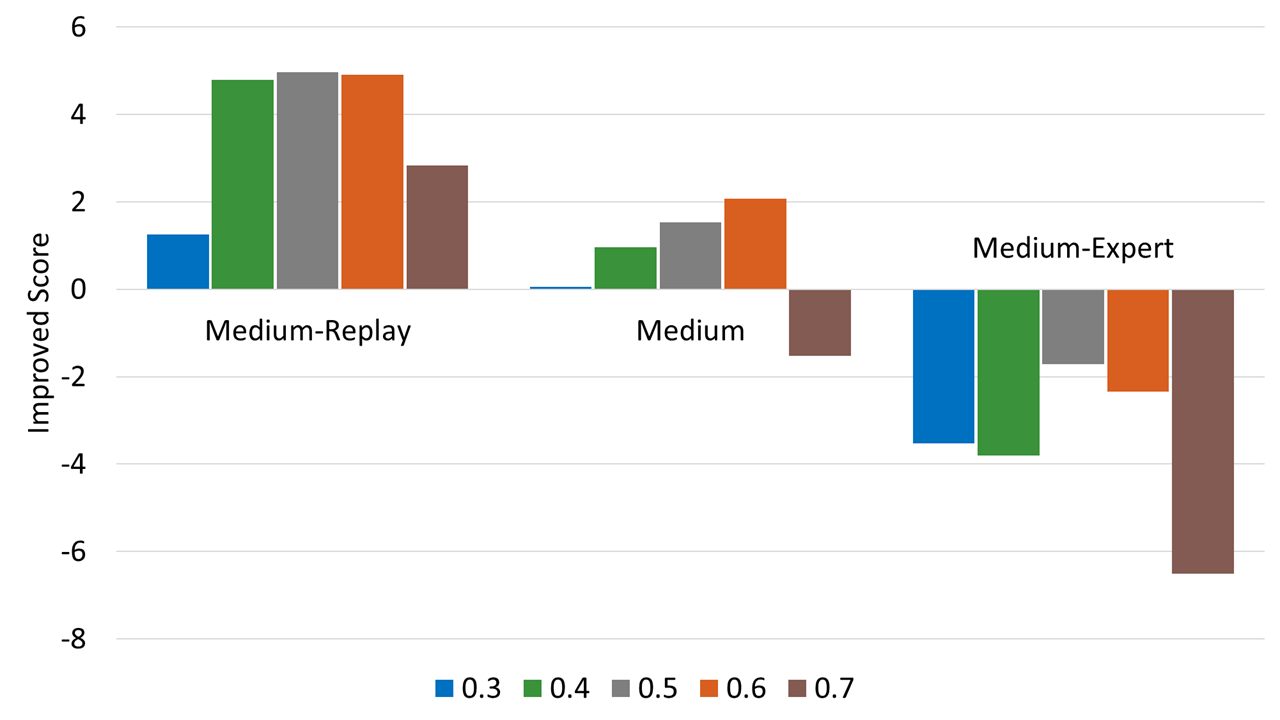}
\caption{Walker2d}
\end{subfigure}
\begin{subfigure}[b]{0.45\textwidth}
\includegraphics[width=\textwidth]{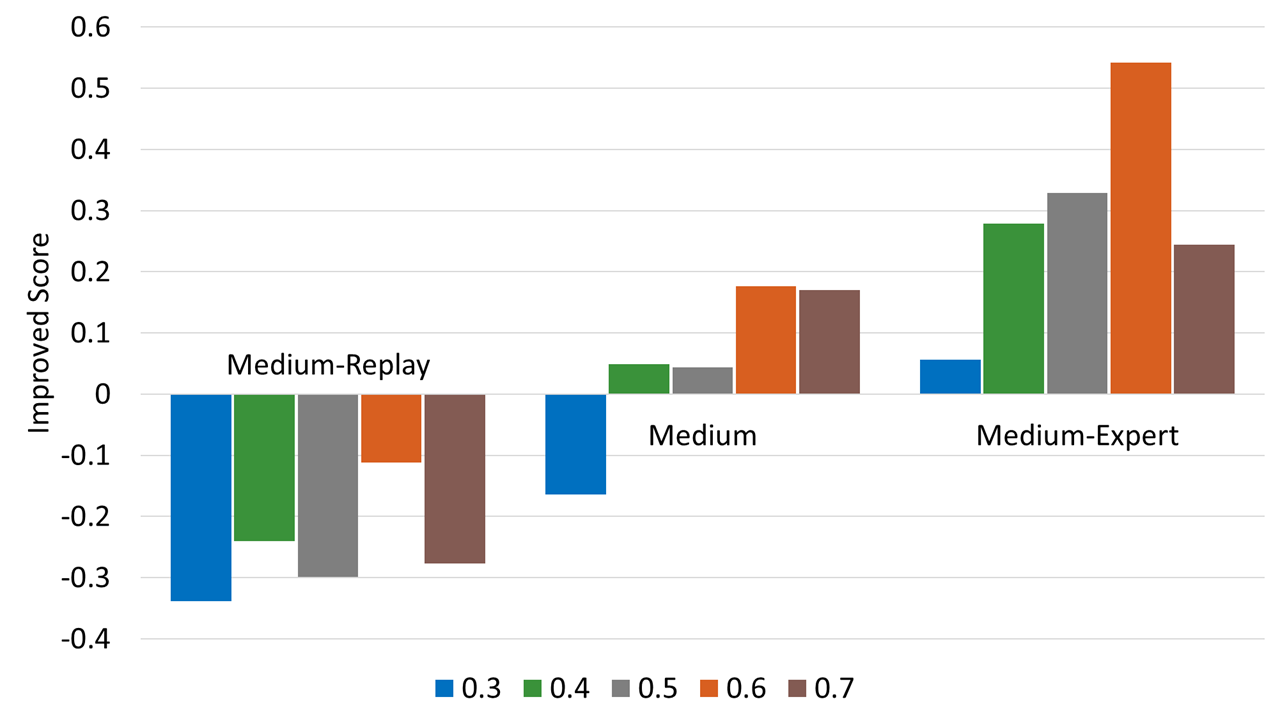}
\caption{Halfcheetah}
\end{subfigure}
\begin{subfigure}[b]{0.45\textwidth}
\includegraphics[width=\textwidth]{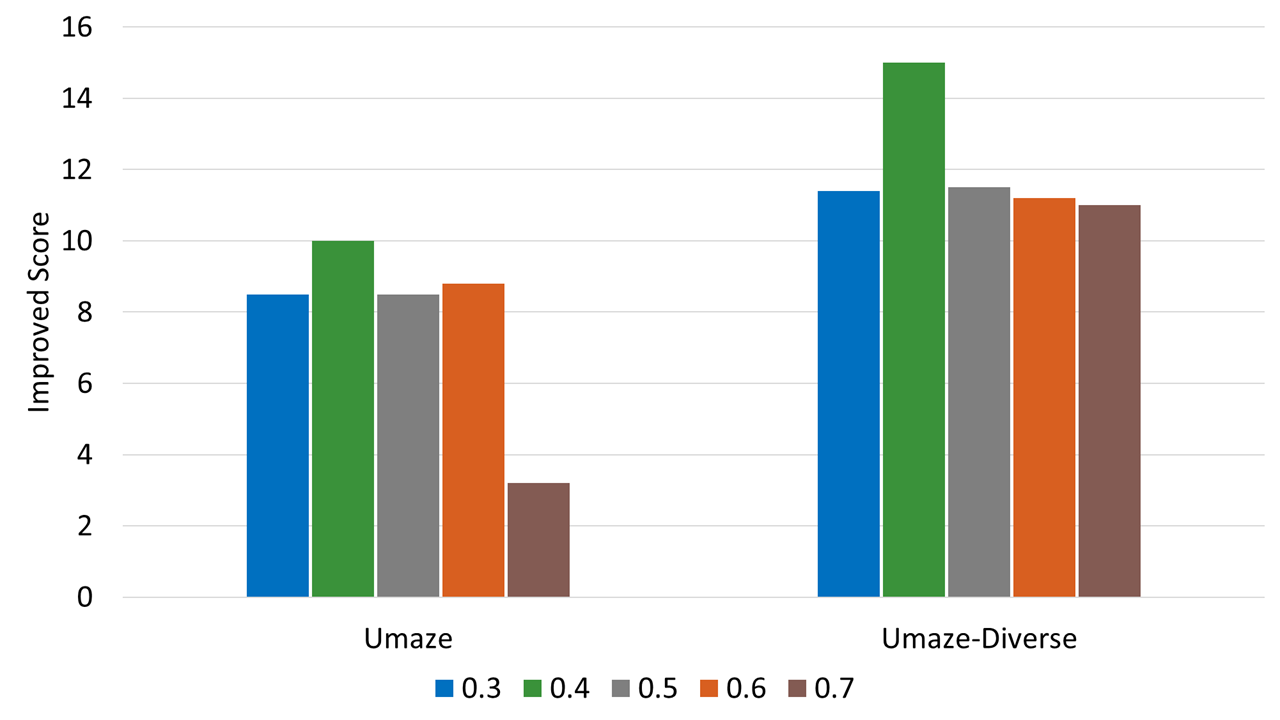}
\caption{Antmaze}
\end{subfigure}
\caption{Detailed ablations of $\tau_c$ in critic training. We show the improved performance of CGDT over DT with different $\tau_c \in \{ 0.3, 0.4, 0.5, 0.6, 0.7 \}$ on different datasets. A large $\tau_c$ indicates a bias over high-quality data, vice versa.}
\label{sup:fig_critic}
\end{figure*}

\begin{figure*}[htbp]
\centering
\begin{subfigure}[b]{0.45\textwidth}
\includegraphics[width=\textwidth]{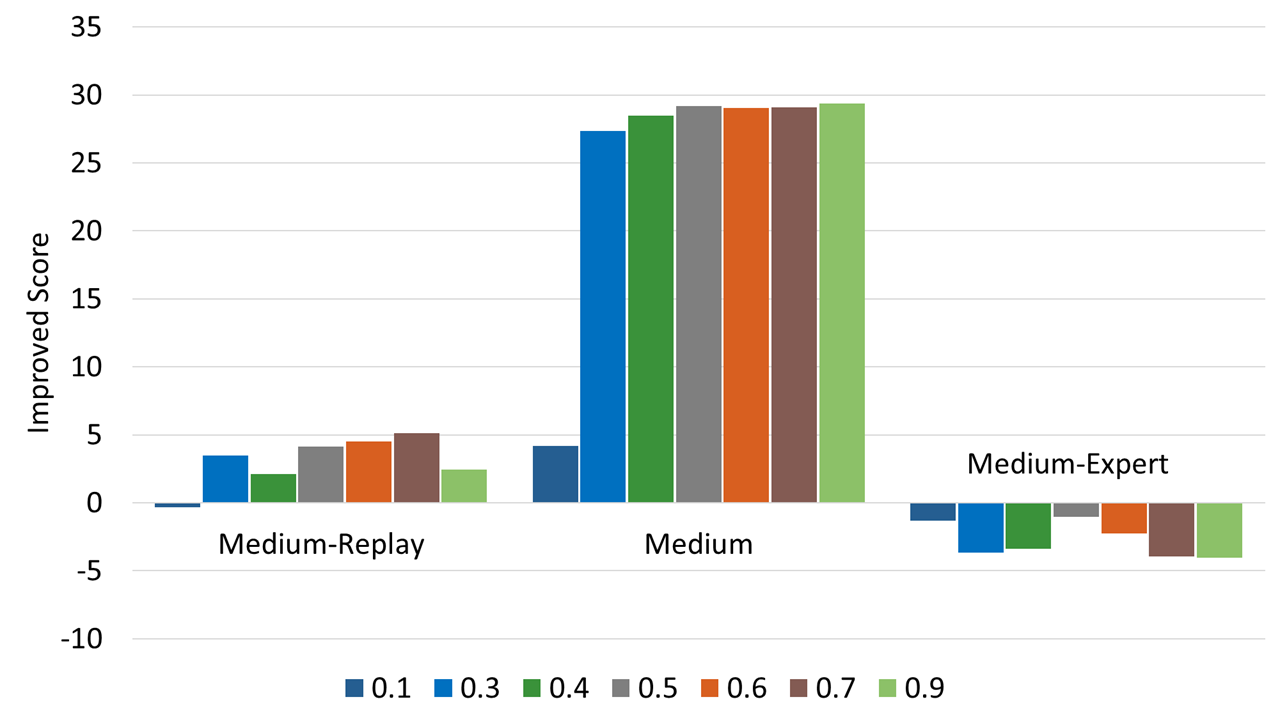}
\caption{Hopper}
\end{subfigure}
\begin{subfigure}[b]{0.45\textwidth}
\includegraphics[width=\textwidth]{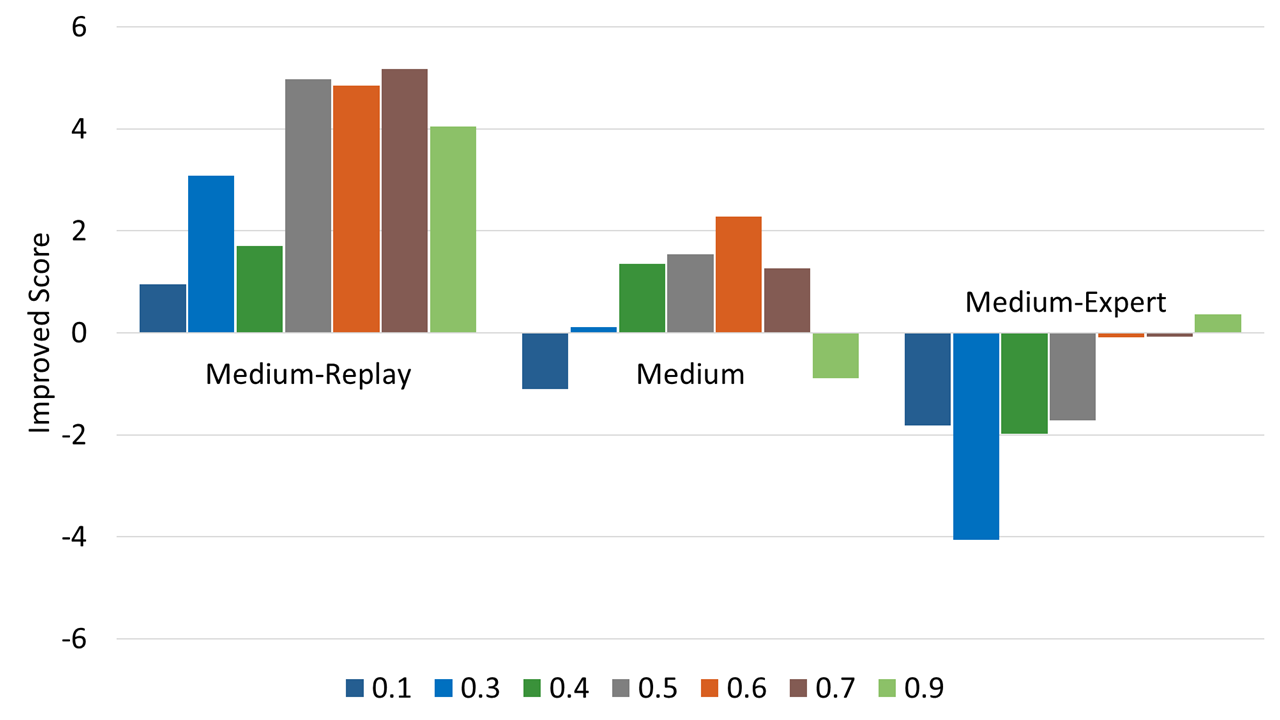}
\caption{Walker2d}
\end{subfigure}
\begin{subfigure}[b]{0.45\textwidth}
\includegraphics[width=\textwidth]{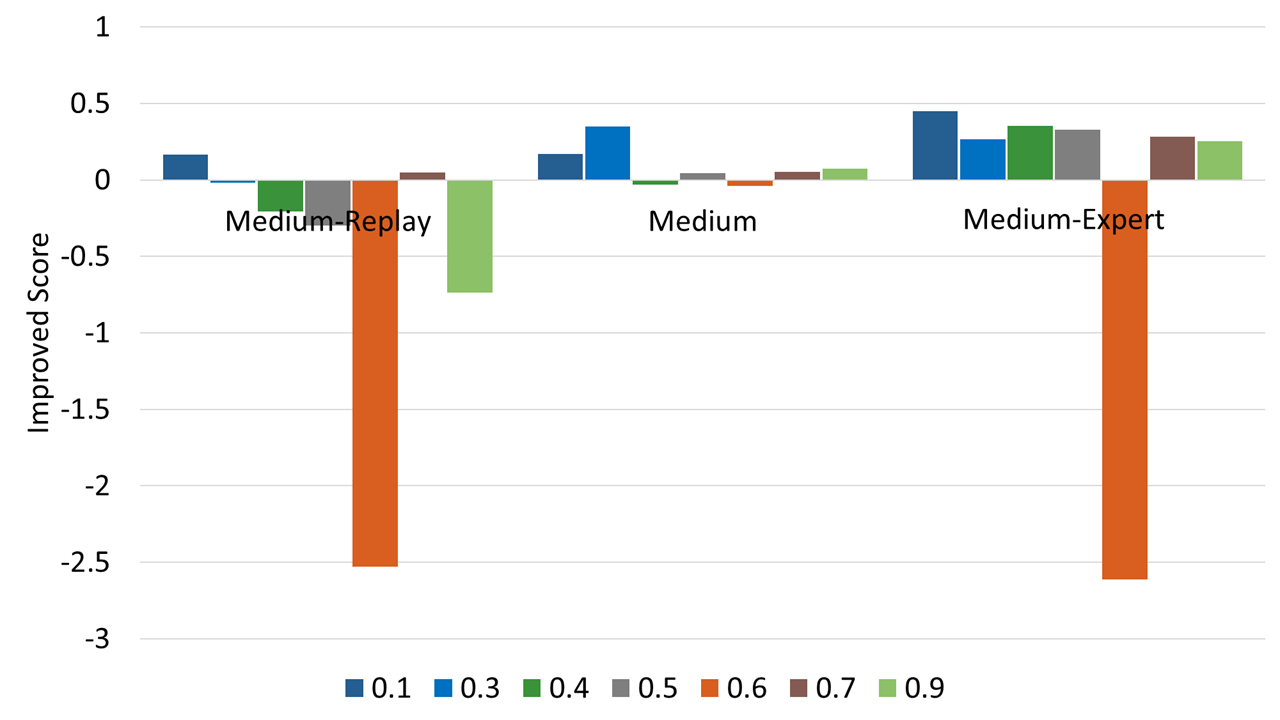}
\caption{Halfcheetah}
\end{subfigure}
\begin{subfigure}[b]{0.45\textwidth}
\includegraphics[width=\textwidth]{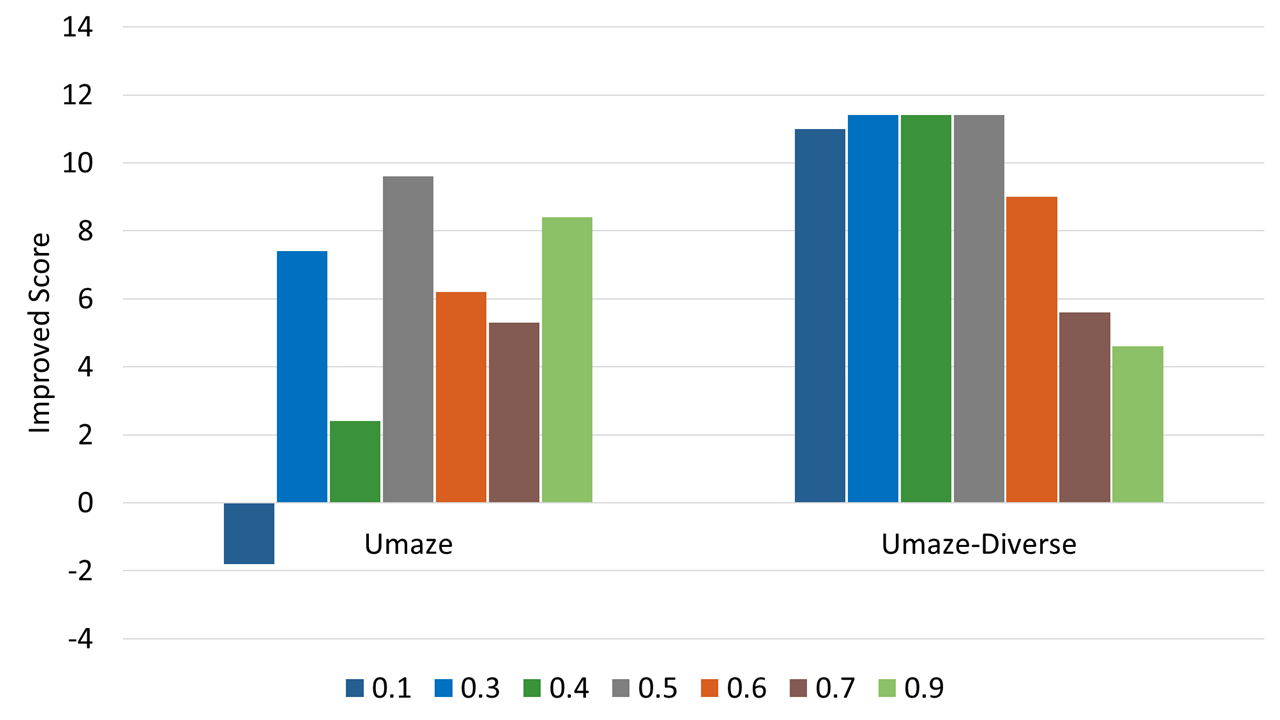}
\caption{Antmaze}
\end{subfigure}
\caption{Detailed ablations of $\tau_p$ in policy training. We show the improved performance of CGDT over DT with different $\tau_p \in \{ 0.1, 0.3, 0.4, 0.5, 0.6, 0.7, 0.9\}$ on different datasets. A large $\tau_p$ favors actions with higher expected returns, vice versa.}
\label{sup:fig_policy}
\end{figure*}

In the main paper, we present the results of an ablation study on $\tau_c$ and $\tau_p$ in the context of summarization across various data qualities. However, it's important to note that the overall performance improvement of CGDT across different data qualities may be heavily influenced by individual tasks that exhibit significant increments or decrements.
To address this, we provide detailed results of the ablation study on $\tau_c$ and $\tau_p$ for each task and dataset in Figure~\ref{sup:fig_critic} and Figure~\ref{sup:fig_policy}.
These figures offer a more comprehensive and granular view of the impact of $\tau_c$ and $\tau_p$ on the performance of CGDT across different tasks and datasets, allowing for a deeper understanding of the specific effects and trends within each context.

Figure~\ref{sup:fig_critic} presents detailed results of the ablation study on $\tau_c$ across different environments. In accord with the observations reported in the main paper, we observe a bias towards larger values of $\tau_c$, indicating a preference for fitting on high-quality data. However, the case of \textit{Antmaze} contradicts this trend as it is a special case with truncated trajectories and zero returns. In \textit{Antmaze}, low returns do not necessarily imply reduced performance, leading to a different pattern of preferences for $\tau_c$ compared to other environments.

Similarly, Figure~\ref{sup:fig_policy} displays the detailed results of the ablation study on $\tau_p$ across different environments. In accord with the main paper's findings, a preference for larger values of $\tau_p$ is observed, indicating a preference for optimistic actions with higher expected returns. However, this preference is reversed in the case of \textit{Halfcheetah}, where a smaller value of $\tau_p$ is preferred. We attribute this anomaly to the significant overestimation observed in the critic training of \textit{Halfcheetah}. The validation errors in critic training imply that the critic trained in \textit{Halfcheetah} tends to overestimate out-of-distribution actions more than in other environments, leading to higher confidence in actions with lower estimated returns. Thus, a smaller $\tau_p$ is preferred to mitigate the side effect caused by overestimation. These findings highlight the need to address the overestimation issue in the critic training process, as it has the potential to significantly influence the performance of CGDT.

\subsection{Additional Results}
We additionally conduct several experiments to investigate the impact of critic guidance in CGDT. Firstly, we explore the effect of training the critic using a percentage $p \in \{0.3, 0.5, 0.7, 0.9, 1.0\}$ of trajectories with the highest returns. Among these percentages, using 90\% of the trajectories results in the highest overall improvement compared to using all trajectories for the \textit{Hopper} and \textit{Walker2d} datasets. This finding suggests that altering the data distribution used to train the critic can be an effective approach to enhance critic guidance.

Secondly, we examine the influence of different scales for replacing $u$ with $(R_t - \mu_t) / \sigma_t$ in the critic guidance learning objective. We observe minor effects when adjusting the balance weight $\alpha$ appropriately. This suggests that tuning $\alpha$ already incorporates the necessary scaling adjustments for $u$.

